\documentclass[a4paper, 10pt, conference]{ieeeconf}      
\usepackage{FG2026}

\FGfinalcopy 

\IEEEoverridecommandlockouts                              
\usepackage{float}
\usepackage{booktabs}
\usepackage{caption}
\usepackage{subcaption}
\usepackage[all]{nowidow}
\usepackage{graphicx}
\usepackage{microtype}
\usepackage{hyperref}
\usepackage{cite}

\usepackage{etoolbox}
\newcounter{BalanceAtReference}
\newcounter{ReferenceIndexForBalancing}
\makeatletter
\def\@balancelastpageonce{%
	\ifnum\value{ReferenceIndexForBalancing}=\value{BalanceAtReference}
	\newpage
	\else
	\relax
	\fi
	\stepcounter{ReferenceIndexForBalancing}
}
\pretocmd{\bibitem}{\@balancelastpageonce}
{} 
{\@latex@error{Patching \bibitem failed}{\@ehd}}
\makeatother

\def\FGPaperID{250} 

\title{\fontsize{15.75pt}{9pt}\selectfont\textbf{
Skarimva: Skeleton-based Action Recognition is a Multi-view Application
}}

\author{\parbox{16cm}{\centering
   {\large Daniel Bermuth, Alexander Poeppel, Wolfgang Reif}\\
   {\normalsize
   Institute for Software and Systems Engineering, University of Augsburg, Germany\\
   \textit{daniel.bermuth@uni-a.de}}}
}

\begin{document}

\maketitle


\begin{abstract}

Human action recognition plays an important role when developing intelligent interactions between humans and machines. 
While there is a lot of active research on improving the machine learning algorithms for skeleton-based action recognition, not much attention has been given to the quality of the input skeleton data itself.
This work demonstrates that by making use of multiple camera views to triangulate more accurate 3D~skeletons, the performance of state-of-the-art action recognition models can be improved significantly.
This suggests that the quality of the input data is currently a limiting factor for the performance of these models. 
Based on these results, it is argued that the cost-benefit ratio of using multiple cameras is very favorable in most practical use-cases, therefore future research in skeleton-based action recognition should consider multi-view applications as the standard setup.

\end{abstract}


\section{Introduction}

To allow a computer system to intelligently react to human actions, it is important that it can recognize these actions reliably.
Skeleton-based action recognition has become a popular approach for this task, because the skeleton data is compact, robust to changes in human appearance and the surroundings, and still preserves the relevant motion information. 

However, most research in this field has focused on improving the machine learning algorithms, while comparatively less attention has been given to the input data itself.
In fact, progress in recent years appears to have plateaued, with only incremental accuracy gains being reported, despite the use of increasingly complex models.
This work demonstrates that by rethinking the input data acquisition process, significant accuracy improvements can be achieved with existing models.
For example, the error rate on the widely used \textit{NTU-RGBD-60} dataset could be reduced by over $50\%$ across different backbones, achieving new state-of-the-art results under standard evaluation protocols.
Therefore, the input data quality seems to have been a limiting factor for the performance of current skeleton-based action recognition models.

Following these results, it is proposed to consider skeleton-based action recognition as a multi-view application, instead of the current single-view standard.
While using multiple cameras increases the system complexity slightly, the setup of additional cameras is relatively simple and worth the small extra effort in most practical applications, as will be discussed later in this work.

To support future research, the complete code is open-sourced
\ifFGfinal
at: \url{https://gitlab.com/Percipiote/}
\else
at: [blinded for review]
\fi

\begin{figure}[t]
    \centering
    \includegraphics[width=0.999\linewidth]{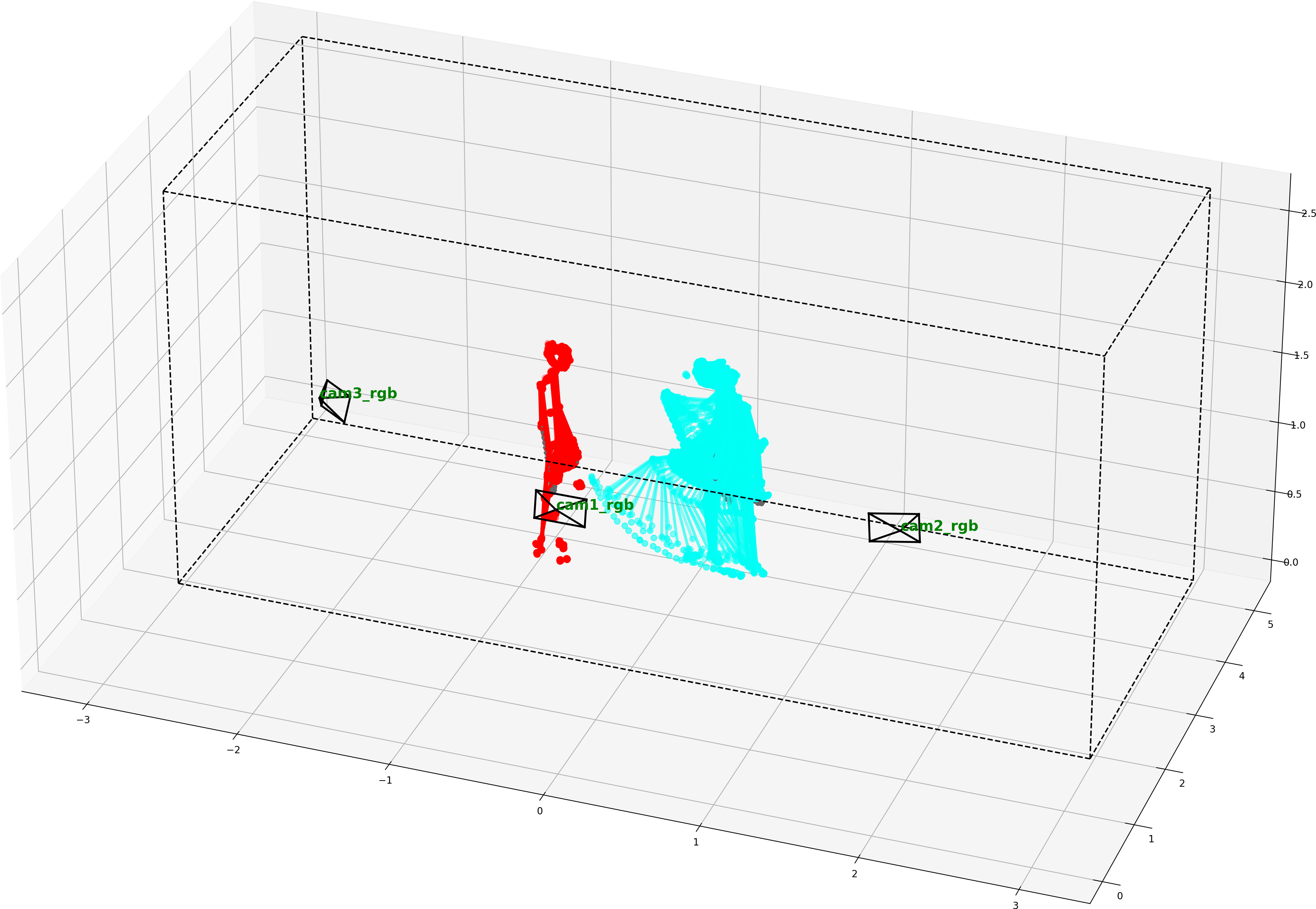}
    \caption{Example of a \textit{kick other person} action with the new multi-view whole-body skeletons.}
    \label{fig:init_example}
    \vspace{-10pt}
\end{figure}

\section{Related Work}

Only a few works have investigated the influence of input skeleton quality so far, which will be the focus in the following.
Since the amount of action-recognition datasets with multi-view image sequences, which are required for this investigation, is very limited, the following experiments are based on the \textit{NTU-RGBD} dataset~\cite{shahroudy2016ntu, liu2019ntu}, which is also the only one of this type that is widely used in the field.
Being one of the most popular action-recognition datasets has furthermore the benefit of allowing an experimental comparison with many previous works.

The original \textit{NTU-RGBD} dataset~\cite{shahroudy2016ntu, liu2019ntu} was created with three non-calibrated \textit{Kinect}~RGB-D cameras, from which each camera's 3D~skeleton estimation was processed independently.
While the skeleton quality was sufficient at the time of release, it still contains a notable amount of poor joint estimations, caused by occlusions, depth ambiguities, and the limited accuracy of the pose estimation algorithms available at the time.
The authors of \textit{PoseConv3D}~\cite{duan2022revisiting} found that using 2D~poses from a more recent pose estimator already improves the action recognition accuracy notably, despite discarding the 3D~distance information.
In \textit{NTU-X}~\cite{trivedi2021ntu}, a different approach was followed, in which new whole-body 2D and 3D~skeletons were created with a single-view 2D-to-3D lifting approach.
This also resulted in accuracy improvements.

Both methods, though, are limited by the single-view setting, which is susceptible to depth ambiguities and occlusion problems.
In contrast, this work explores a multi-view triangulation approach to even further improve the skeleton quality, which leads to significantly better action recognition results.
A more detailed comparison with these previous works is provided later in the experiments section.


\section{Improving Pose Estimates}

One of the main problems that prevents directly using a multi-view approach on the \textit{NTU-RGBD} dataset is that the camera calibrations and synchronizations are not available.
Therefore, a reconstruction procedure was developed to recover intrinsic and extrinsic camera parameters by minimizing multi-view skeleton reprojection and pose alignment errors (using the original camera-aligned skeletons) with iterative outlier removal, as well as to estimate temporal alignment between the video streams.
Additional details about this procedure are provided in the appendix.
The source-code was integrated into the \textit{skelda} library~\cite{bermuth2024voxelkeypointfusion}.

\vspace{3pt}
To calculate new skeleton poses, the state-of-the-art 3D~multi-view multi-person pose estimator \textit{RapidPoseTriangulation}~\cite{bermuth2025rapidposetriangulation} can now be used.
It was shown to generalize well between different camera setups and datasets, and is one of the very few which also supports whole-body pose estimation with face and finger keypoints.
It also has the benefit that it does not require re-training for each camera setup, which is often the case in learned multi-view approaches~\cite{bermuth2025rapidposetriangulation}.
The pose estimation, with about $130$~FPS on an Nvidia~RTX4080, runs faster than the camera frame-rate of $30$~FPS.

\vspace{3pt}
To create motion sequences again, the skeletons are matched to persons over time using a simple dataset-optimized tracking approach.
Specifically, a new skeleton is assigned to an existing track if the average joint distance is below a certain threshold.
If no matching track is found, a new track is initialized. After all skeletons have been assigned to tracks, short tracks below a certain length are removed as false detections.
If two tracks do not overlap in time, they are merged into one.
Since this dataset has at most two persons in a scene, only the two longest tracks are kept and additional tracks are discarded.


\section{Results}

To evaluate the benefits of the new multi-view skeletons, three recent skeleton-based action recognition models were trained and evaluated on the \textit{NTU-RGBD-60}~\cite{shahroudy2016ntu} and \textit{NTU-RGBD-120}~\cite{liu2019ntu} datasets.
The models used are \textit{MSG3D}~\cite{liu2020disentangling}, \textit{DG-STGCN}~\cite{duan2022dg}, and \textit{ProtoGCN}~\cite{liu2025revealing}, which are all graph convolutional network (GCN) based models that have shown strong performance on these datasets.
For the first two models, the implementations from \textit{PySKL}~\cite{duan2022pyskl} were used.
In general, only a few modifications were necessary to train the models with the new skeletons, mostly related to the different joint counts and types.
The training hyperparameters were kept identical to the original implementations as far as possible, to ensure comparability.
As augmentations, only random rotations around the z-axis (which are now easy with the world-aligned skeletons) and size scaling were used.

\subsection{Improvements with new whole-body poses}

Table~\ref{tab:ntu_wb} demonstrates that using the new multi-view whole-body skeletons leads to significant accuracy improvements for all three models on both dataset splits.
Note that in contrast to the commonly used multi-stream ensembling approach, in this experiment all input modalities (\textit{j+b+jm+bm}) are concatenated and fed into a single model, which reduces the training effort notably.
Further, only a single sequence is sampled for each test sequence to speed up testing.

\begin{table}[H]
    \centering
    \begin{tabular}{|l||c||c|c|c|c|c|}
        \toprule
        Method        & original & wb25 & wb137 & wb69 & wb31 & body \\
        \midrule
        \textit{MSG3D}    & 90.8 & 96.5 & 94.0 & 96.3 & 96.4 & 94.0 \\
        \textit{DG-STGCN} & 92.0 & 96.7 & 94.4 & 96.5 & 96.9 & 94.6 \\
        \textit{ProtoGCN} & 91.8 & 97.1 & 94.3 & 95.6 & 96.6 & 95.5 \\
        \midrule
        \textit{MSG3D}    & 87.0 & 93.6 & 90.4 & 93.3 & 93.5 & 88.2 \\
        \textit{DG-STGCN} & 88.3 & 94.6 & 92.7 & 94.9 & 94.1 & 89.7 \\
        \textit{ProtoGCN} & 88.7 & 94.8 & 91.6 & 94.4 & 94.3 & 89.9 \\
        \bottomrule
    \end{tabular}
    \caption{Results with the new whole-body poses on \textit{NTU-RGBD-60} and \textit{NTU-RGBD-120}. The split \textit{wb25} has the same keypoints as the original one, \textit{wb137} has all whole-body keypoints, \textit{wb69} has no additional face keypoints, and \textit{wb31} only adds the other three fingertips.}
    \label{tab:ntu_wb}
\end{table}

An interesting observation is that adding more finger and face keypoints does not necessarily improve the accuracy further.
This could be an indication that the models start to overfit on keypoints that are not particularly relevant for the action classes.
This effect was also observed in \textit{NTU-X}~\cite{trivedi2021ntu}.
Additionally, the models become significantly slower with more input joints due to the strong increase in graph edges.
These findings raise the question for future work of how the models can be designed to use the whole-body keypoints more efficiently and focus more on the relevant keypoints automatically themselves.

\subsection{Comparison with other skeleton recalculations}

Besides the approach followed in this work, two other methods for improving the skeletons of \textit{NTU-RGBD} have been evaluated already.
As described previously, \textit{PoseConv3D}~\cite{duan2022revisiting} has introduced higher-quality 2D body-only skeletons from a more recent 2D estimator, and \textit{NTU-X}~\cite{trivedi2021ntu} new whole-body skeletons with a single-view 2D-to-3D lifting approach.
The results in Table~\ref{tab:ntux} show that the multi-view triangulation approach proposed in this work outperforms both previous methods by a significant margin.

\begin{table}[H]
    \centering
    \begin{tabular}{|l|c|c|c|c|}
        \toprule
        Method & body & wb25 & body+fingers & body+hand+face \\
        \midrule
        \textit{original}~\cite{liu2020disentangling} & -    & 91.5 & - & - \\
        \textit{PoseConv3D}~\cite{duan2022revisiting} & 92.5 & -  & - & - \\
        \textit{NTU60-X}~\cite{trivedi2021ntu}        & 91.3 & - & 91.8 & 91.1 \\
        \textit{Skarimva}                             & \textbf{94.0} & \textbf{96.5} & \textbf{96.3} & \textbf{94.0} \\
        \midrule
        \textit{original}~\cite{liu2020disentangling} & - & 86.9 & - & - \\
        \textit{PoseConv3D}~\cite{duan2022revisiting} & 87.2 & -  & - & - \\
        \textit{NTU120-X}~\cite{trivedi2021ntu}       & 84.5 & - & 87.1 & - \\
        \textit{Skarimva}                             & \textbf{88.2} & \textbf{93.6} & \textbf{93.3} & \textbf{90.4} \\
        \bottomrule
    \end{tabular}
    \caption{Comparison with other skeleton representations using the \textit{MSG3D} model. Note that there are small differences in the joint types and ensembling methods, but they are not expected to significantly impact the results, as explained in more detail in the text.}
    \label{tab:ntux}
\end{table}

\vspace{6pt}
Note that there exist some slight training differences, but they are unlikely to significantly affect the results.
The \textit{NTU-X} \textit{body} set includes $6$ foot keypoints, whereas the one from \textit{PoseConv3D} and \textit{Skarimva} only use the standard \textit{COCO}~\cite{lin2014microsoft} body joints.
The \textit{body+fingers} set is basically the same, except that \textit{Skarimva} uses $2$ additional joints derived from the body joints.
The face joints of \textit{NTU-X} do not contain keypoints for the cheeks, so their \textit{body+fingers+face} set has $118$ joints instead of the $137$ in \textit{Skarimva}.
Regarding model ensembling, the original results of \textit{MSG3D} use a two-stream approach with joint and bone input modalities (\textit{j+b}), in which each output of the two models is ensembled by averaging the two prediction distributions.
The papers of \textit{NTU-X}~\cite{trivedi2021ntu} and \textit{PoseConv3D}~\cite{duan2022revisiting} do not explain their ensembling method, but from the source-code it seems they use the same.
The results with \textit{Skarimva} do not use ensembling, but a single model with all four input modalities (\textit{j+b+jm+bm}), as in Table~\ref{tab:ntu_wb}.
This way only one model has to be trained, which notably reduces training effort, and as shown in Table~\ref{tab:inpmod60}~and~\ref{tab:inpmod120} in the appendix, the accuracy stays on a similar level.

\vspace{6pt}
While the results of \textit{PoseConv3D}~\cite{duan2022revisiting} with the better body-only skeletons show that improving the quality of the 2D~pose estimates already is beneficial, the authors also noticed a larger drop in performance when projecting the original 3D poses back to 2D (\cite{duan2022revisiting},\,Table\,16), caused by the loss of helpful 3D information.
This can also be seen when comparing the accuracy to the new 3D body-only poses of this work, which is clearly higher (while both methods used different 2D pose estimators, their performance is on a similar level, so the influence from this should be minimal).

The results of \textit{NTU-X}~\cite{trivedi2021ntu} show that adding fingers to the skeletons is important to improve the average accuracy, but their lifting-based 3D reconstruction method is notably inferior to the multi-view approach proposed in this work. 
Another experiment in \textit{PoseConv3D} also showed that lifting itself does not help (\cite{duan2022revisiting},\,Table\,13).
An explanation for this effect could be that the action recognition model already receives the pose input over time, so its input is similar to that of the lifting models.
Therefore, if doing so would be helpful, the action model could learn to perform the lifting internally.
Under this assumption an additional lifting model would not add any new information, but instead could introduce additional errors from the lifting process.
The only benefit of an external lifting model could be that it can be pre-trained with general human motions on a larger dataset, because no action labels are required.
However, such a pre-training would be possible for the multi-view pose estimation approaches as well.
In fact, most of the lifting datasets are created from multi-view datasets already.
In theory, depth triangulation is possible in a single-viewpoint video over time as well, using invariant distances (like limb lengths) moving in front of the camera, but multi-view triangulation is much easier and more computationally efficient.
Using multiple views also has the important advantage that (self-)\,occlusions can be partially avoided, whereas a video model would need to infer occluded keypoints from context, which is susceptible to errors.

\newpage
\subsection{With multi-stream ensembling}

The common evaluation procedure of previous works uses multi-stream ensembling of multiple input modalities, combined with creating multiple randomized samples from a single test sequence, to improve the accuracy further. 
Note that since this procedure is computationally very intensive, it is not suited for real-time applications.
The results with this evaluation method are given in Table~\ref{tab:ntu_both} for completeness, and set new state-of-the-art scores.
The model performance on the single modalities is provided in the appendix.

\begin{table}[H]
    \centering
    \begin{tabular}{|l|c|}
        \toprule
        Method & xsub \\
        \midrule
        \textit{MSG3D}~\cite{liu2020disentangling}         \hspace{20pt}'2020 & 91.5 \\
        \textit{CTR-GCN}~\cite{chen2021channel}            \hspace{16pt}'2021 & 92.4 \\
        \textit{ST-GCN++}~\cite{duan2022pyskl}             \hspace{08pt}'2022 & 92.6 \\
        \textit{MotionBert}~\cite{zhu2023motionbert}       \hspace{09pt}'2023 & 93.0 \\
        \textit{InfoGCN}~\cite{chi2022infogcn}             \hspace{20pt}'2022 & 93.0 \\
        \textit{BlockGCN}~\cite{zhou2024blockgcn}          \hspace{11pt}'2024 & 93.1 \\
        \textit{MMP-ST}~\cite{zhou2025multi}               \hspace{16pt}'2025 & 93.1 \\
        \textit{LGS-Net}~\cite{pan2025language}            \hspace{18pt}'2025 & 93.2 \\
        \textit{DG-STGCN}~\cite{duan2022dg}                \hspace{11pt}'2022 & 93.2 \\
        \textit{HD-GCN}~\cite{lee2023hierarchically}       \hspace{15pt}'2023 & 93.4 \\
        \textit{SkateFormer}~\cite{do2024skateformer}      \hspace{09pt}'2024 & 93.5 \\
        \textit{LA-GCN}~\cite{xu2023language}              \hspace{17pt}'2023 & 93.5 \\
        \textit{TDSN-GCN}~\cite{liu2025tdsn}               \hspace{07pt}'2025 & 93.6 \\
        \textit{MSA-GCN}~\cite{alowonou2024msa}            \hspace{15pt}'2024 & 93.6 \\
        \textit{DE-GCN}~\cite{myung2024degcn}              \hspace{16pt}'2024 & 93.6 \\
        \textit{JMDA}~\cite{xiang2025joint}                \hspace{26pt}'2025 & 93.7 \\
        \textit{Hyper-GCN}~\cite{zhou2025adaptive}         \hspace{07pt}'2025 & 93.7 \\
        \textit{Shap-Mix}~\cite{zhang2024shap}             \hspace{15pt}'2024 & 93.7 \\
        \textit{ProtoGCN}~\cite{liu2025revealing}          \hspace{11pt}'2025 & 93.8 \\
        \textit{PoseC3D}~\cite{duan2022revisiting}         \hspace{16pt}'2022 & 94.1 \\
        \textit{Hulk}~\cite{wang2025hulk}                  \hspace{30pt}'2025 & 94.3 \\
        \textit{SkeletonAgent}~\cite{liu2025skeletonagent} \hspace{00pt}'2025 & 94.5 \\
        \textit{3Mformer}~\cite{wang20233mformer}          \hspace{13pt}'2023 & 94.8 \\
        \textit{LLM-AR}~\cite{qu2024llms}                  \hspace{18pt}'2024 & 95.0 \\
        \textit{POTR}~\cite{cao2025reenvisioning}          \hspace{30pt}'2025 & 95.3 \\
        \midrule
        \textit{ProtoGCN+Skarimva} & \textbf{97.5} \\
        \bottomrule
    \end{tabular}
    \hspace{8pt}
    \begin{tabular}{|l|c|}
        \toprule
        Method & xsub \\
        \midrule
        \textit{MSG3D}~\cite{liu2020disentangling} & 86.9  \\
        \textit{PoseC3D}~\cite{duan2022revisiting} & 86.9  \\
        \textit{ST-GCN++}~\cite{duan2022pyskl} & 88.6  \\
        \textit{LLM-AR}~\cite{qu2024llms} & 88.7  \\
        \textit{CTR-GCN}~\cite{chen2021channel} & 88.9  \\
        \textit{DG-STGCN}~\cite{duan2022dg} & 89.6 \\
        \textit{SkateFormer}~\cite{do2024skateformer} & 89.8 \\
        \textit{InfoGCN}~\cite{chi2022infogcn} & 89.8  \\
        \textit{HD-GCN}~\cite{lee2023hierarchically} & 90.1  \\
        \textit{MMP-ST}~\cite{zhou2025multi} & 90.2 \\
        \textit{BlockGCN}~\cite{zhou2024blockgcn} & 90.3 \\
        \textit{Shap-Mix}~\cite{zhang2024shap} & 90.4 \\
        \textit{MSA-GCN}~\cite{alowonou2024msa} & 90.6 \\
        \textit{LA-GCN}~\cite{xu2023language} & 90.7 \\
        \textit{JMDA}~\cite{xiang2025joint} & 90.9 \\
        \textit{Hyper-GCN}~\cite{zhou2025adaptive} & 90.9 \\
        \textit{ProtoGCN}~\cite{liu2025revealing} & 90.9 \\
        \textit{DE-GCN}~\cite{myung2024degcn}     & 91.0 \\
        \textit{TDSN-GCN}~\cite{liu2025tdsn} & 91.1 \\
        \textit{POTR}~\cite{cao2025reenvisioning} & 91.1 \\
        \textit{SkeletonAgent}~\cite{liu2025skeletonagent} & 91.7 \\
        \textit{3Mformer}~\cite{wang20233mformer} & 92.0 \\
        \midrule
        \textit{ProtoGCN+Skarimva} & \textbf{95.4} \\
        \bottomrule
    \end{tabular}
    \caption{Results on \textit{NTU-RGBD-60} and \textit{NTU-RGBD-120}.}
    \label{tab:ntu_both}
\end{table}

\subsection{Few-shot learning}

In real-world applications, it is often unlikely to have large amounts of labeled training data (here around $200$ training examples per class), because labeling is time-consuming and costly.
Therefore, the ability to classify new actions with only a few examples is an important feature for action recognition models.
Thus, the following experiments evaluate few-shot learning capabilities with the new skeleton poses.

Because the source-code of the top three previous works was either incomplete or not available, the currently used classification models were extended instead.
In particular, their default linear-layer classification head was replaced by a combined contrastive embedding and classification head, and the data sampling process was modified to support the new training scenario.
The idea of the embeddings is to map samples of the same class closer together in the embedding space, while pushing samples of different classes further apart. 
In evaluation, when new classes are introduced, their embeddings can be calculated from a single example, and then new samples can be classified by calculating the shortest distance to the example embeddings.

Table~\ref{tab:1shot} shows that the current state-of-the-art performance for one-shot learning can be improved notably with the new skeletons as well.

\begin{table}[H]
    \centering
    \begin{tabular}{|l|c|}
        \toprule
        Method &  120 \\
        \midrule
        \textit{APSR}~\cite{liu2019ntu} & 45.3 \\
        \textit{TCN OneShot}~\cite{sabater2021one} &  46.5 \\
        \textit{SL-DML}~\cite{memmesheimer2021sl} &  49.6 \\
        \textit{Skeleton-DML}~\cite{memmesheimer2022skeleton} &  54.2 \\
        \textit{MotionBert}~\cite{zhu2023motionbert} &  61.0 \\
        \textit{Koopman}~\cite{wang2023neural} &  68.1 \\
        \textit{M+C-scale}~\cite{yang2024one} &  68.7 \\
        \textit{SkeletonX}~\cite{zhang2025skeletonx} &  69.1 \\
        \midrule
        \textit{ProtoGCNce+Skarimva} & \textbf{76.0} \\
        \bottomrule
    \end{tabular}
    \caption{One-shot transfer classification on \textit{NTU-RGBD}.}
    \label{tab:1shot}
\end{table}

Using only one labeled example is considered unreasonable though, because in practical applications it should be possible to provide a few more examples with low effort, especially if accuracy benefits from it.
Therefore, Table~\ref{tab:5shot} investigates the results with five examples per action class, a scenario that leads to notably better performance.

\begin{table}[H]
    \centering
    \begin{tabular}{|l|c|}
        \toprule
        Method &  120 \\
        \midrule
        nearest neighbor    &  84.7 \\
        5-nearest neighbor  & 84.9 \\
        prototype matching  & 84.9 \\
        \bottomrule
    \end{tabular}
    \caption{Few-shot transfer with five examples per action.}
    \label{tab:5shot}
\end{table}

\subsection{Real-time Applications}

Real-time capability is an important factor for many applications. However, as it is outside the scope of this section and unrelated to the new skeleton poses, some further notes on this topic are provided in the appendix.


\section{Discussion}

The title of this work frames skeleton-based action recognition as a multi-view problem, which is discussed in more detail in this section.

\vspace{6pt}
The experimental results clearly showed that improving the 3D~skeleton quality by multi-view triangulation has led to significant performance gains for state-of-the-art skeleton-based action recognition models. 
This suggests that the quality of the input data was a limiting factor for the performance of these models. 

Using a multi-view approach is a straightforward way to improve the 3D~pose quality, as depth ambiguities and occlusions can be reduced by observing the subject from different angles.
A similar principle underlies binocular vision in animals, where multiple viewpoints improve depth perception and robustness.

\vspace{6pt}
From a practical perspective, the main drawback of a multi-view setup is the need for additional cameras.
However, the added system complexity is often moderate in relevant application scenarios.

In professional setups, like sports analytics, surveillance, or robotics, the effort of installing another camera should be negligible compared to the overall system setup, and often multiple cameras are used already for other purposes anyway.
In consumer applications, those in which reliable recognition accuracy has some importance, a simple multi-camera setup can be realized as well with limited effort.
Basically, non-experienced users could buy two or three inexpensive USB~cameras, plug them into a computer, and mount them in a static position.
Then they could calibrate them by printing out a chessboard pattern, or displaying one on their smartphone screens, and moving it around in front of the cameras, while they are guided by a software tool.
For mobile applications, like action recognition on smartphones, many devices already have multiple cameras built-in nowadays, which could be used for multi-view triangulation if their relative distance is large enough, otherwise an external clip-on camera could be added instead.

While accurate camera calibration and the synchronization of their image streams is beneficial, especially in professional setups, it is not strictly necessary.
The observed gains in this work were achieved with rather rough camera calibrations and without synchronization at all.
So an inexpensive home setup as explained before will be able to benefit from the additional views as well.

Regarding the additional computational cost, because now multiple image streams have to be processed, this is not a big issue either.
As mentioned earlier, \textit{RapidPoseTriangulation} can process multiple images on consumer-grade hardware in real-time already, and faster than most consumer-grade cameras will provide them.

\vspace{6pt}
In summary, using multiple views for skeleton-based action recognition is a simple and effective way to notably improve the recognition accuracy, offering a favorable cost-benefit ratio in many practical applications.


\section{Conclusion}

This work has shown that improving the input data quality is an effective way to boost the performance of skeleton-based action recognition models.
In particular, it has been demonstrated that leveraging multiple camera views to triangulate 3D~skeletons leads to substantially better skeleton poses, and in the following, to large accuracy gains.
Across multiple settings, the accuracy error could be reduced by over~$50\%$ compared to the original skeletons, achieving new state-of-the-art results on the popular \textit{NTU-RGBD} dataset.

Based on these findings, it can be concluded that skeleton-based action recognition should be a multi-view application, even though most research in this field uses single-view skeletons so far.
Since the cost-benefit ratio of using multiple cameras is very favorable in many scenarios, it is recommended to use multiple cameras if possible for practical applications.

To support future research, for example towards even better camera calibration, or a more effective integration of all whole-body joints into the action recognition models, code and models used in this work are made publicly available.



\newpage

{
   \small
   \bibliographystyle{ieee}
   \bibliography{main}
}

\clearpage
\appendix

\subsection{Calibration Details}

As mentioned in the main text, the calibration process can be split into three steps:

\begin{itemize}
    \vspace{3pt}
    \item Estimate the camera intrinsics
    \item Estimate the camera extrinsics
    \item Synchronize the image streams
    \vspace{3pt}
\end{itemize}

To estimate the intrinsics, the provided original 2D and 3D skeletons can be used.
The 3D skeletons are projected into the image plane using an initial guess for the intrinsic camera matrix and distortion coefficients.
Then, the re-projection error between the projected 3D joints and the provided 2D joints is calculated and used to minimize the initial intrinsic parameters. 
To stabilize the optimization, the initial intrinsic parameters were calculated as an average of the \textit{Kinect}~cameras used in the \textit{Panoptic} dataset~\cite{joo2015panoptic}, which uses the same camera model.

\vspace{3pt}
To estimate the extrinsics, a similar approach is taken.
The provided 3D skeletons in camera coordinates should align with each other in world coordinates (except for the pose estimation errors). 
An initial guess for the extrinsic parameters is created by placing the cameras in a circle around the room origin, looking towards the center, using the height and distance mentioned in the dataset paper. 
The first central camera serves as static reference, and all camera-originated skeletons are transformed into world coordinates using the initial extrinsic parameters.
Then an error is calculated between the first camera's skeleton and the other cameras' skeletons, which is used to optimize the extrinsic parameters of the two side-view cameras.
After one round of optimization, $30\%$ of the frames with the highest error are removed as outliers, and the optimization is repeated. This is done two times in total.
To skip the additional matching complexity, only sequences with a single subject, and where the number of skeleton frames is equal for all three cameras, are used for \mbox{the calibration process}.

\begin{figure}[H]
    \centering
    \includegraphics[width=0.97\linewidth]{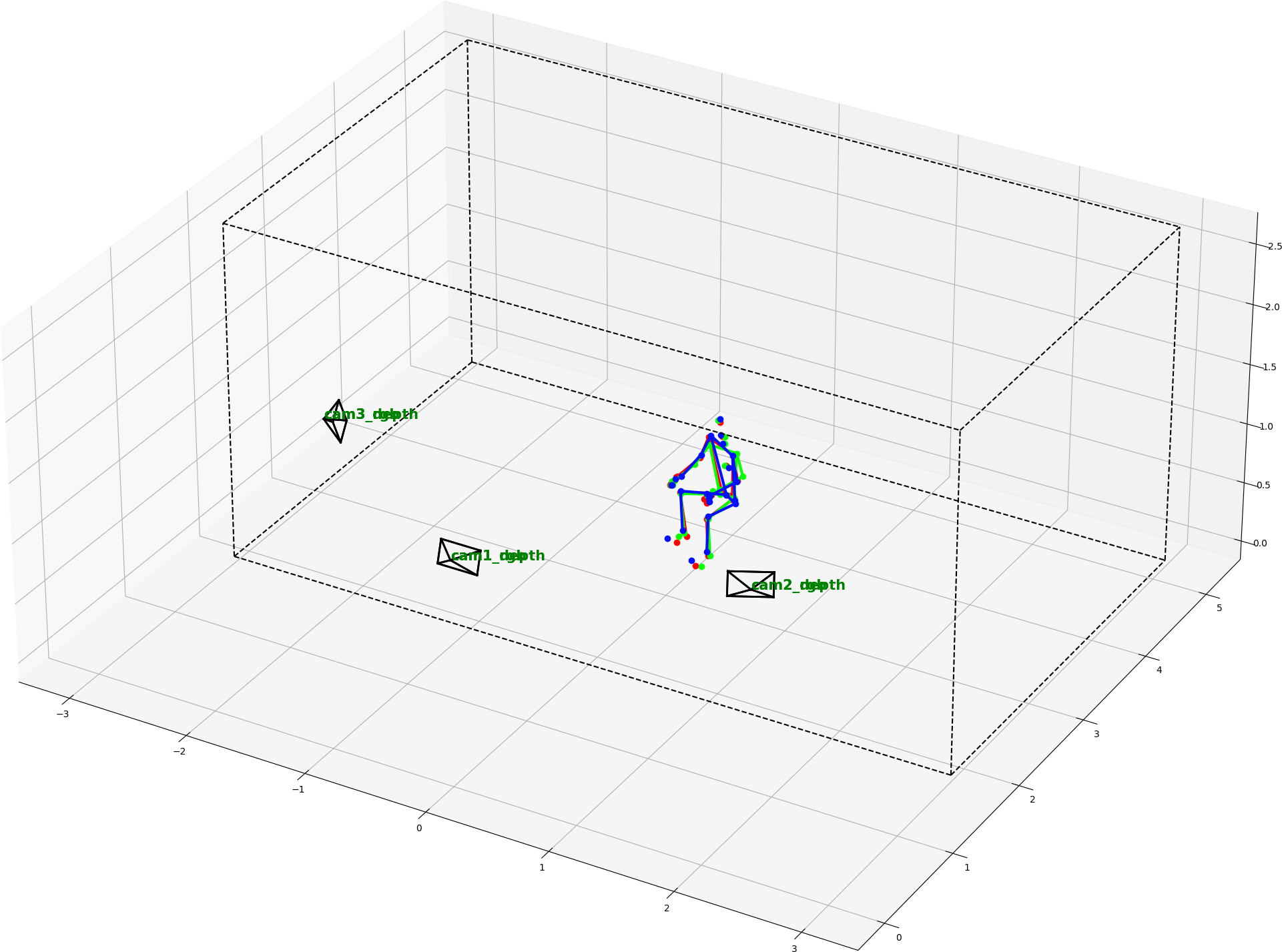}
    \caption{Extrinsic calibration by overlapping skeletons from the different views. The overlap cannot be perfect due to errors in the original pose estimations.}
    \label{fig:calib_example}
\end{figure}

\vspace{3pt}
To synchronize the images, a relatively simple approach is used.
The original videos are split into image frames, and because the videos often do not have the same length and also slightly varying frame-rate, the average frame count is calculated, and the image frames of each camera are uniformly sampled by skipping or duplicating frames to match this average count.
This way, the images are roughly aligned in time, although not perfectly synchronized.

\vspace{3pt}
A task for future work could be to further improve the calibration and synchronization, by also using image-space features, like objects in the background or the motion of the person in the different views.
Such a method could then use the calibration parameters from this work to initialize their optimization algorithm. 
The current method calculates an average calibration for each setup and performer, but sometimes the cameras have slightly moved between different actions, or in the duration of one recording, and such movements are not taken into account yet, and can result in slight calibration offsets.
But for now the results are already good enough to demonstrate the benefits of multi-view skeleton estimation.

\subsection{Real-time Applications}

Many applications require responding to human actions with only a small delay.
Therefore, the prediction speed of the models is of interest as well.
As can be seen in Table~\ref{tab:specs}, the three tested models are rather small and can predict actions faster than most cameras deliver images, making them real-time capable.

\begin{table}[H]
    \centering
    \begin{tabular}{|l|c|c|}
        \toprule
        Method & Size & FPS \\
        \midrule
        \textit{MSG3D}    & 3.14M & 189 \\
        \textit{DG-STGCN} & 2.04M & 140 \\
        \textit{ProtoGCN} & 4.46M & 150 \\
        \bottomrule
    \end{tabular}
    \caption{Model size and inference speed. Tested on a single Nvidia-RTX4080 with a batch-size of $1$.}
    \label{tab:specs}
\end{table}

By default, the input of the models is a complete action sequence.
From those, a fixed number of images (here $100$) is randomly sampled and fed into the model.
For this approach the start and end of an action have to be known beforehand, but this is not very realistic for streamed camera inputs in a live setting.
Therefore, to evaluate such a setting, the following experiment samples a fixed-length continuous sequence at a random starting position from the action sample instead.
This would be similar to a windowed streaming approach.
The results in Table~\ref{tab:seqsamp} show that this leads to a slight performance decrease, growing with smaller window sizes, but the accuracy is still reasonably high.

\begin{table}[H]
    \centering
    \begin{tabular}{|l|c|c|c|c|}
        \toprule
        Method & 3s & 2s & 1s \\
        \midrule
        \textit{DG-STGCN} & 95.7 & 94.7 & 92.8 \\
        \textit{ProtoGCN} & 95.6 & 94.7 & 92.8 \\
        \bottomrule
    \end{tabular}
    \caption{Comparison of the effect of different input durations on \textit{NTU-RGBD-60}, using a single random continuous subsequence of each sample.}
    \label{tab:seqsamp}
\end{table}

\subsection{Ensembling Modalities}

For completeness, the performance results on the single input modalities are provided in the following.
It can be seen that the combined \textit{j+b+jm+bm} input is better than one modality alone, but if runtime is of no concern, training the models separately and fusing the outputs afterwards improves the results even more.

\begin{table}[H]
    \centering
    \begin{tabular}{|c|c|c|c|c|}
        \toprule
        Joint & Bone & J-Motion & B-Motion \\
        \midrule
        95.6 & 96.4 & 93.3 & 93.8 \\
        \bottomrule
    \end{tabular}
    \begin{tabular}{|c|}
        \toprule
        J+B+JM+BM \\
        \midrule
        97.1 \\
        \bottomrule
    \end{tabular}
    \begin{tabular}{|c|c|}
        \toprule
        1J+1B-Fusion  & 2J+2B+1JM+1BM-Fusion \\
        \midrule
        97.1 & 97.2 \\
        \bottomrule
    \end{tabular}
    \\
    \begin{tabular}{|c|c|}
        \toprule
        K  & K-Motion \\
        \midrule
        96.6 & 94.5 \\
        \bottomrule
    \end{tabular}
    \begin{tabular}{|c|}
        \toprule
        2J+2B+2K+1JM+1BM+1KM-Fusion \\
        \midrule
        97.5 \\
        \bottomrule
    \end{tabular}
    \caption{Comparison of input modalities and output fusions, using \textit{ProtoGCN} on \textit{NTU-RGBD-60-xsub}.}
    \label{tab:inpmod60}
\end{table}

\begin{table}[H]
    \centering
    \begin{tabular}{|c|c|c|c|c|}
        \toprule
        Joint & Bone & J-Motion & B-Motion \\
        \midrule
        91.9 & 93.4 & 89.5 & 90.1 \\
        \bottomrule
    \end{tabular}
    \begin{tabular}{|c|}
        \toprule
        J+B+JM+BM \\
        \midrule
        94.8 \\
        \bottomrule
    \end{tabular}
    \begin{tabular}{|c|c|}
        \toprule
        1J+1B-Fusion  & 2J+2B+1JM+1BM-Fusion \\
        \midrule
        94.7 & 95.0 \\
        \bottomrule
    \end{tabular}
    \\
    \begin{tabular}{|c|c|}
        \toprule
        K  & K-Motion \\
        \midrule
        93.9 & 89.8 \\
        \bottomrule
    \end{tabular}
    \begin{tabular}{|c|}
        \toprule
        2J+2B+2K+1JM+1BM+1KM-Fusion \\
        \midrule
        95.4 \\
        \bottomrule
    \end{tabular}
    \caption{Comparison of input modalities and output fusions, using \textit{ProtoGCN} on \textit{NTU-RGBD-120-xsub}.}
    \label{tab:inpmod120}
\end{table}

Note that this evaluation concept reduces the runtime significantly, by $10\times$ for multi-sampling each test sequence and by another $6\times$ due to the different modalities. 
So the \textit{ProtoGCN} model can only process around $3$ instead of $150$ samples per second, notably reducing its real-time capability.

\vspace{3pt}
Because the new skeletons are in world-coordinates, cross-setup generalization is expected to be relatively straightforward for the models.
The \textit{NTU-RGBD-120} dataset also has a \textit{xset} split that was intended to test this. 
But it has the problem that some new setups also contain new subjects as well, so it does not specifically test the cross-setup generalization alone.
This makes it difficult to isolate only the effect of setup changes, so the accuracy results provide no directly usable insights, and for those reasons no evaluation was done on the \textit{xset} split. 
Besides that, the \textit{xsub} split already includes setup switches as well.

\vspace{3pt}
Instead of just using differences in the inputs, another option is to also use differences in the model architectures, by ensembling the different models as well.
Table~\ref{tab:model_fusion} shows that this can result in further improvements, but as before, at the cost of real-time performance.

\begin{table}[H]
    \centering
    \begin{tabular}{|l|c|c|}
        \toprule
        Method & 60 & 120 \\
        \midrule
        \textit{MSG3D} + \textit{DG-STGCN} + \textit{ProtoGCN} & 97.9 & 96.0 \\
        \bottomrule
    \end{tabular}
    \caption{Ensembling different model architectures.}
    \label{tab:model_fusion}
\end{table}

\subsection{About objects and image inputs}

While this work does not explore it, the benefit of multi-view inputs should also be applicable to approaches that use color images for action recognition as well. 
Especially with the \textit{NTU-RGBD} dataset, the additional images contain valuable information about objects that are being manipulated in some actions, which is not available in the skeleton data.
Distinguishing whether a person is drinking a cup of water or eating a hamburger is quite difficult with skeleton data only, but should be much easier when the image data is available as well.
The confusion matrices also show a higher error rate at many actions which involve objects.

A possible downside of such a combined model could be that, because the recognition model will get more input data, it will lead to lower few-shot performance, because the model focuses more on the object appearance than the motion patterns.
But this is just speculation at this point, and should be investigated in future work.
A further downside would be the increased computational cost, which is an important factor in real-time applications.
For example in the robotic application of \textit{Tutabo}~\cite{bermuth2025tutabo}, the pose estimations are calculated on edge devices, because sending the image streams to a central computer would already be too time-consuming.
So a combined image+skeleton model would also need to run on the edge device, which could notably increase its computational requirements.

A possible option in such a scenario, and one that would also fit the skeleton-based input concept, could be to use object bounding-box detections, triangulate them similarly to the skeletons, and use them as additional input to the action recognition model.
This also would be a privacy-friendly approach for distributed or non-local applications, because only the sparse skeleton and object position data would need to be transmitted, instead of the full images.
But again, this will be left for future work.

\subsection{Confusion Matrices}

\begin{figure*}
    \centering
    \includegraphics[width=0.999\linewidth]{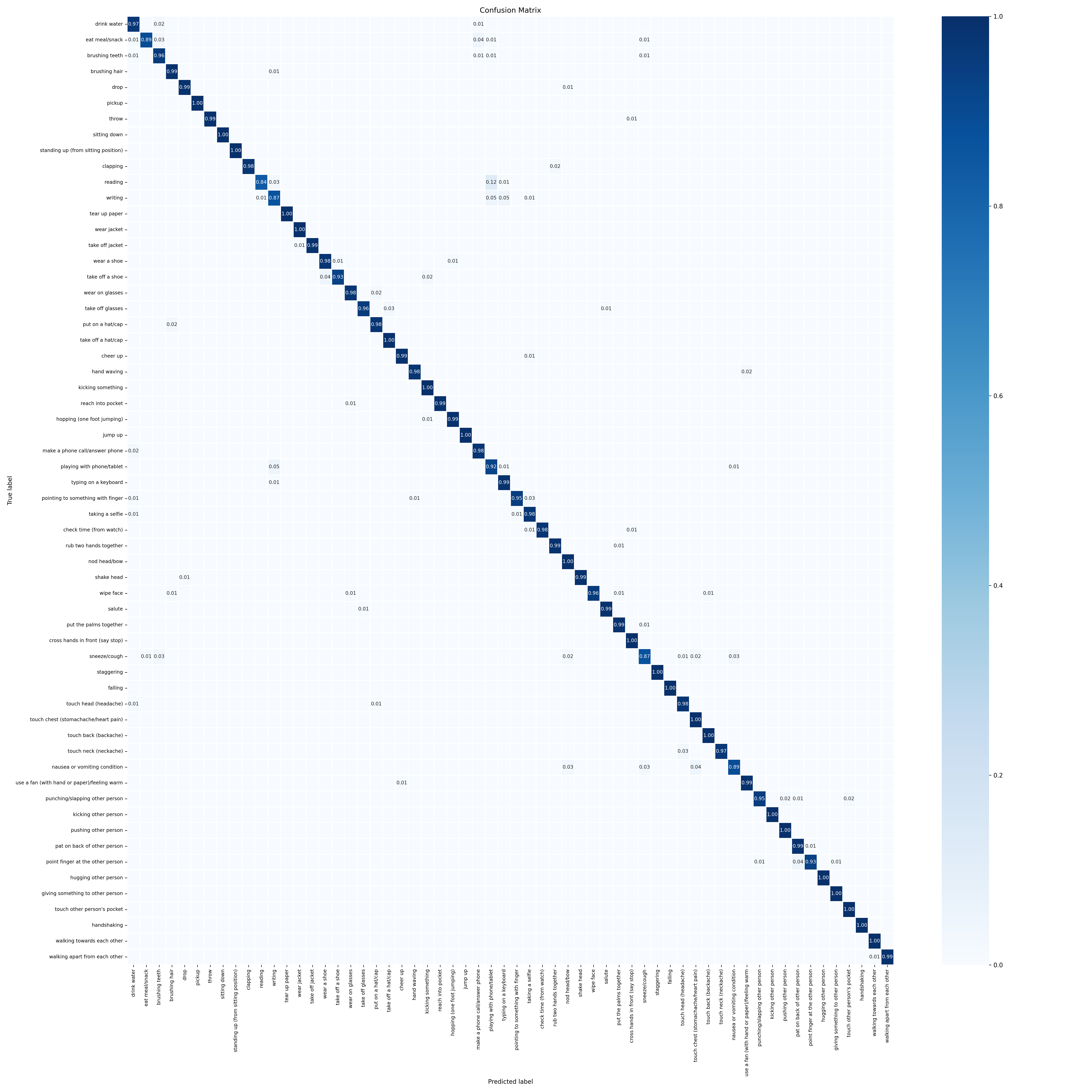}
    \caption{Confusion matrix of the ensembled \textit{ProtoGCN} model on \textit{NTU-RGBD-60-xsub}.}
    \label{fig:cm60}
\end{figure*}

\begin{figure*}
    \centering
    \includegraphics[width=0.999\linewidth]{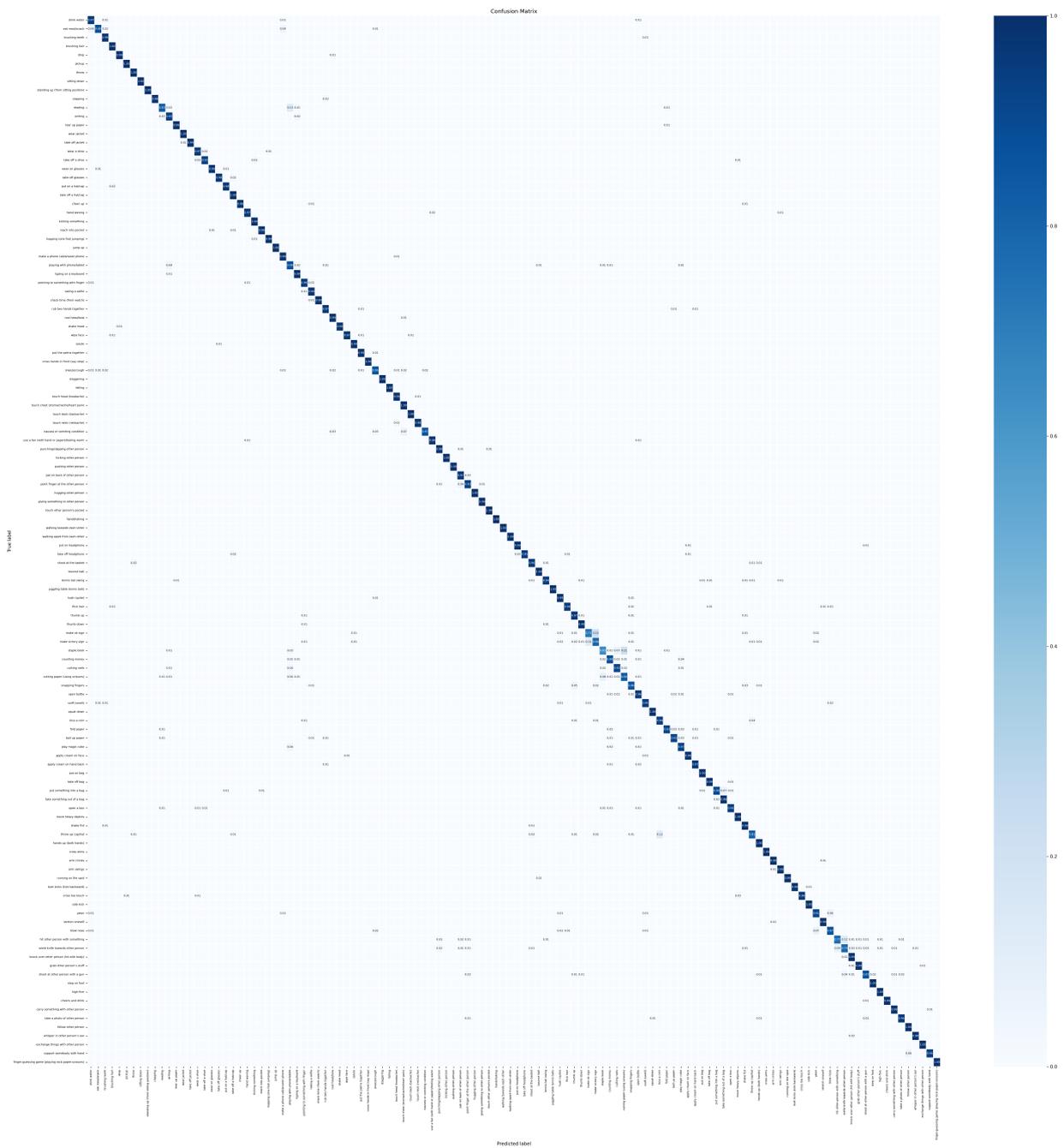}
    \caption{Confusion matrix of the ensembled \textit{ProtoGCN} model on \textit{NTU-RGBD-120-xsub}.}
    \label{fig:cm120}
\end{figure*}

The confusion matrices in Figure~\ref{fig:cm60} and Figure~\ref{fig:cm120} show that most actions are classified very well, and confusions often occur between similar ones. 
In many cases the (here not existing) knowledge about objects could help the model, for example to distinguish between \textit{eat snack} and \textit{drink water} or between \textit{reading} and \textit{play with tablet}. 

\end{document}